\pdfoutput=1

\documentclass[11pt]{article}

\usepackage{emnlp2021}

\usepackage{times}
\usepackage{latexsym}

\usepackage[T1]{fontenc}

\usepackage[utf8]{inputenc}

\usepackage{microtype}

\usepackage{booktabs}
\usepackage{graphicx}

\usepackage{layouts} 

\usepackage{todonotes}
\newcommand{\dgaddy}[1]{{}}
\newcommand{\kuza}[1]{{}}
\newcommand{\mpavan}[1]{{}}
\newcommand{\pkolhar}[1]{{}}

\title{Overcoming Conflicting Data when Updating a Neural Semantic Parser}

\author{David Gaddy \\
  University of California, Berkeley\thanks{\hspace{1em}Work performed during an internship at Google.} \\
  \texttt{dgaddy@berkeley.edu} \AND
Alex Kouzemtchenko, Pavankumar Reddy Muddireddy, Prateek Kolhar, \\
and \textbf{Rushin Shah} \\
  Google Assistant \\
  \texttt{\{kuza,mpavan,pkolhar,rushinshah\}@google.com}}

\date{}

\begin{document}
\maketitle
\begin{abstract}

In this paper, we explore how to use a small amount of new data to update a task-oriented semantic parsing model when the desired output for some examples has changed.  When making updates in this way, one potential problem that arises is the presence of conflicting data, or out-of-date labels in the original training set.  To evaluate the impact of this understudied problem, we propose an experimental setup for simulating changes to a neural semantic parser.  We show that the presence of conflicting data greatly hinders learning of an update, then explore several methods to mitigate its effect.  Our multi-task and data selection methods lead to large improvements in model accuracy compared to a naive data-mixing strategy, and our best method closes 86\% of the accuracy gap between this baseline and an oracle upper bound.

\end{abstract}

\section{Introduction}

Most work in semantic parsing (and NLP in general) considers a scenario where the desired outputs of a model are static and can be specified with a large, fixed dataset.
However, when deploying a semantic parsing model in a real world virtual assistant, it is often necessary to update a model to support new features or to enable improvements in downstream processing.
For example, creators of a media assistant\kuza{media example seems unclear since it's never talked about again} may want to add new commands specialized towards new media types like podcasts, or those of a navigation assistant might like to add a new feature to allow users to specify roads to avoid.
Such changes require that the structures output by the model are updated, either by introducing a new intent (a representation of a desired action) or re-configuring arguments of existing intents.
In this work, we investigate the best way to make updates to the intents and arguments of a task-oriented neural semantic parsing model.

To make an update, new data annotations must be collected to specify the new form that is desired for model outputs\mpavan{and the existing data must be updated to conform to the new outputs}\mpavan{does calling them outputs instead of output schema or something like that introduce confusion?}.
Because changes can be quite frequent, we would like to be able to collect a small amount of data for each update (on the order of 50 examples) and merge the new information with a much larger existing dataset.
Naively, we might hope to simply combine new data in with the old and train on the combination.
However, this approach has the problem that some of the older data with out-of-date labels may conflict with the new labeling.
These conflicts occur whenever inputs that would be affected by a change appear in the original dataset.
For example, when introducing a new intent for podcasts, the original dataset may have included podcast examples labeled with a more generic media intent or a label indicating that the feature is unsupported.
When introducing a new argument, say `roads to avoid', there may be instances in the original dataset that should have this argument labeled but do not because they were annotated before the argument was introduced.
This conflicting data can confuse the model and cause it to predict labels as they were before an update rather than after.
Unfortunately, this problem of conflicting data during model updates is understudied in the academic literature.

To enable exploration of this problem on publicly available datasets, we propose a method to easily create simulated updates with conflicts (Section~\ref{sec:synthetic-changes}) and release our update data.
The idea behind our method is to form updates in the reverse direction, relabeling instances of a particular intent or argument to simulate out-of-date labels.
Using our proposed setup, we demonstrate how conflicting data greatly hinders learning of updates (Section~\ref{sec:conflicting-test}).

In addition, we explore several methods for mitigating the negative effects of conflicting data by modifying how we combine the old and the new data (Section~\ref{sec:methods}).
One approach is to keep the old and new datasets separate, but to share information indirectly with either fine-tuning or multi-task learning.
Another approach is to explicitly filter data that is likely to conflict using a learned classifier.
Each of these methods substantially improves performance compared to naively mixing in new data, establishing strong baselines for future work on this task.

In summary, the contributions of this work are 1)~establishing an experimental setup to test updates with conflicting data, 2)~demonstrating that conflicting data leads to large losses in model performance if left unmitigated, and 3)~exploring several possible approaches to mitigate this problem.

\section{Related Work}

There has been a substantial amount of prior work on making updates to neural models 
\cite{xiao2014error,rusu2016progressive, li2016learning,kirkpatrick2017overcoming,castro2018end}, demonstrating a recognition in the community that the ability to update models is important.
However, most of these works consider a setting where none of the original data conflicts with the new labels.
Thus these works, and the general continual learning and class-incremental learning literature, assume that the space of inputs affected by a change does not appear at all in the original data.
In many scenarios, this assumption does not hold because the original dataset will aim to cover the full distribution of inputs a model might encounter.
Because of the non-conflicting assumption, this body of prior work focuses on other questions such as what can be done when the original data is no longer available.

One paper that does consider updates with label conflicts is \citet{chen2019transfer}.
Although they do not intentionally set out to study conflicting-data updates, in their NER task locations where new labels apply are tagged with special ``outside'' labels prior to the update, which cause conflicts with the new labels.
While their work avoids the conflicting data problem by considering a setting where the original data is no longer available, our experiments show that it can be advantageous to instead keep the original data around and use more direct methods to avoid problems of conflicting data.

Our work also has parallels to the concept drift literature \cite{tsymbal2004problem,lu2018learning}.
However, concept drift work focuses on unintentional changes to a natural data distribution over time, while our work is concerned with intentional re-structuring of the output representation space, leading to very a different setup and choice of methods.
In particular, that work operates over a stream of examples where changes often occur gradually, does not generally include structured prediction tasks, and does not allow the practitioner to introduce additional structure on the types of annotation given (as we do in Section~\ref{sec:setup-updates}).

Finally, our work relates to work on training with noisy labels \cite{sukhbaatar2014training,Veit2017LearningFN,Jiang2018MentorNetLD}, since the incorrect labels from conflicting data could be viewed as a type of noise.
However, it is important to evaluate the problem of conflict-causing updates separately from other noisy label problems because the distribution of incorrect labels due to an update will be very different from most other sources of label noise.
While not all noisy-label methods can directly apply to our task (as many are designed for classification as opposed to structured prediction) and may not take full advantage of the additional structure of our problem, we believe this line of work can still serve as a source of inspiration for future exploration on our task.

\section{Task Setup}
\label{sec:setup}

\subsection{Preliminaries - Task-oriented Semantic Parsing}

The experiments in this paper focus on a semantic parsing task where the goal is to generate a tree structure conditioned on an input sentence.
We use the task formulation and data from \citet{gupta-etal-2018-semantic-parsing} as the basis of our setup.
Output trees are made up of intents and arguments (aka slots), where intents come from a fixed inventory of labels, and arguments consist of an argument-type label along with a value.  Argument values may either be free-form text selected from the input sentence, or a nested intent to form a hierarchical structure.
In this work, we represent these trees with a linearized form using nested brackets, which allows for the use of standard sequence-to-sequence models (see Section~\ref{sec:base-model} for details of our base model and Figure~\ref{fig:examples} below for some example inputs and outputs).

\subsection{Data Updates}
\label{sec:setup-updates}

In this paper, we focus on the task of making a single update to the intent and argument structure output by a model, where the update is specified by collecting a small amount of additional data.\footnote{While handling a stream of updates may pose additional challenges, we leave an investigation of that scenario to future work.  What qualifies as a single update is somewhat open to interpretation, but our methods are not overly sensitive to how it is defined.}
Accordingly, our task setup expects two sets of data: a large amount of data from before a change, which we will call the \textbf{V1} (version one) set, and a small amount of data from after a change, which we call the \textbf{V2} set.  The V1 set represents the current state of the system with any data collected in the past, while the new V2 set is collected specifically for the purpose of introducing a particular update.  For the purposes of testing methods in this paper, we will form these two sets synthetically, as described in Section~\ref{sec:synthetic-changes} below.

\newcommand{\partition}[1]{\emph{#1}}

Because the V2 set is gathered explicitly for the purpose of introducing a particular update, we would like the input distribution of this data to be targeted rather than uniformly covering the full input space.
Ideally, a substantial portion of this data should be examples whose labels will actually change after the update.  We will call this portion of data the \textbf{\partition{changed}} set.
However, it is also useful to specify some examples that are not affected by a change, so that we can accurately determine the scope of a change.  We will call inputs whose label would be the same under the new and old label scheme \textbf{\partition{unchanged}} examples.  We will include a set of \partition{unchanged} examples in the V2 data to show that these labels have been confirmed under the V2 labeling scheme.

Many types of changes that we care about only apply to examples labeled in a particular way in the original V1 data.  For example, a new argument can often only be used for specific intents, or a new intent may only apply to examples previously labeled as unsupported.  By taking advantage of this information, we may be able to avoid some unwanted side effects of our model updates.  To help us handle this information, we define a third category of examples: \textbf{\partition{trivially-unchanged}}.  The \partition{trivially-unchanged} partition contains all examples which we can determine to be unaffected based solely on the original V1 labels and some simple hand-defined rule like a list of affected intents.
By identifying these examples, we can directly include them in the updated training set without causing label conflicts.
In the remainder of this paper, we reserve the term \partition{unchanged} to refer specifically to unchanged examples that do not fall into the \partition{trivially-unchanged} category.
Thus, the \partition{unchanged} partition represents the remaining hard examples that are difficult to distinguish from the \partition{changed} set in the V1 data.
See Figure~\ref{fig:examples} for examples of how our three data partitions apply to particular updates.

In some instances, it is also useful to talk about the data in the original V1 set in terms of the three partitions (\partition{changed}, \partition{unchanged}, and \partition{trivially-unchanged}).
In this case, these labels refer to whether an example \emph{would} change if we had gathered new labels for them.
In actuality, the \partition{changed} subset will have out-of-date labels in V1, and we call these examples \emph{conflicting data}.
Our experiments show that conflicting data causes substantial problems when learning an update (see Section~\ref{sec:conflicting-test} and \ref{sec:evaluation}), but unfortunately, the examples that make up the conflicting set cannot be easily identified in the V1 data.

\definecolor{dkgreen}{HTML}{38761d}
\begin{figure*}[]
    \centering
    \small
    \begin{tabular}{lll}
        \toprule \addlinespace[0.7em]
        \multicolumn{3}{c}{\textbf{New intent from related}} \vspace{.5em} \\
        \multicolumn{3}{l}{\textbf{Changed}} \vspace{.5em} \\
        \hspace{4em} & \multicolumn{2}{l}{Query: \emph{Where is there construction on the highway?}} \vspace{.5em} \\
        & \hspace{2em} V1 Label: & \hspace{2em} V2 Label: \\
        & \texttt{(IN:GET\_INFO\_ROAD\_CONDITION} & \texttt{\textcolor{dkgreen}{\textbf{(IN:GET\_INFO\_TRAFFIC}}} \\
        &  \hspace{1em} \texttt{(SL:LOCATION "the highway" ) )} & \hspace{1em} \texttt{(SL:LOCATION "the highway" ) )} \vspace{.5em} \\
        \multicolumn{3}{l}{\textbf{Unchanged}} \vspace{.5em} \\
        & Query: \emph{Are roads icy?} \vspace{.5em} \\
        & \hspace{2em} V1 Label: & \hspace{2em} V2 Label: \\
        & \texttt{(IN:GET\_INFO\_ROAD\_CONDITION} & \texttt{(IN:GET\_INFO\_ROAD\_CONDITION} \\
        & \hspace{1em} \texttt{(SL:ROAD\_CONDITION "icy" ) )} & \hspace{1em} \texttt{(SL:ROAD\_CONDITION "icy" ) )} \vspace{.5em} \\
        \multicolumn{3}{l}{\textbf{Trivially-unchanged}} \vspace{.5em} \\
        & \multicolumn{2}{l}{Examples not labeled with \texttt{IN:GET\_INFO\_ROAD\_CONDITION} in V1 are trivially-unchanged.} \\
        
        \\ \midrule \addlinespace[0.7em]
        \multicolumn{3}{c}{\textbf{New intent from unsupported}} \vspace{.5em} \\
        \multicolumn{3}{l}{\textbf{Changed}} \vspace{.5em} \\
        & \multicolumn{2}{l}{Query: \emph{If I leave right now, can I get to New York City before one o'clock PM?}} \vspace{.5em} \\
        & \hspace{2em} V1 Label: & \hspace{2em} V2 Label: \\
        & \texttt{(IN:UNSUPPORTED\_NAVIGATION )} & \texttt{\textcolor{dkgreen}{\textbf{(IN:GET\_ESTIMATED\_ARRIVAL}}} \\
        &  & \hspace{1em} \texttt{(SL:DATE\_TIME\_DEPARTURE "right now" )} \\
        &  & \hspace{1em} \texttt{(SL:DESTINATION "New York City" ) )} \vspace{.5em} \\
        \multicolumn{3}{l}{\textbf{Unchanged}} \vspace{.5em} \\
        & \multicolumn{2}{l}{Query: \emph{What major city has the worst traffic?}} \vspace{.5em} \\
        & \hspace{2em} V1 Label: & \hspace{2em} V2 Label: \\
        & \texttt{(IN:UNSUPPORTED\_NAVIGATION )} & \texttt{(IN:UNSUPPORTED\_NAVIGATION )} \vspace{.5em} \\
        \multicolumn{3}{l}{\textbf{Trivially-unchanged}} \vspace{.5em} \\
        & \multicolumn{2}{l}{Examples not labeled with \texttt{IN:UNSUPPORTED\_NAVIGATION} in V1 are trivially-unchanged.} \\
        
        \\ \midrule \addlinespace[0.7em]
        \multicolumn{3}{c}{\textbf{New argument}} \vspace{.5em} \\
        \multicolumn{3}{l}{\textbf{Changed}} \vspace{.5em} \\
        & \multicolumn{2}{l}{Query: \emph{Which route to work has less traffic?}} \vspace{.5em} \\
        & \hspace{2em} V1 Label: & \hspace{2em} V2 Label: \\
        & \texttt{(IN:GET\_DIRECTIONS} & \texttt{(IN:GET\_DIRECTIONS} \\
        & \hspace{1em} \texttt{(SL:DESTINATION "work" ) )} & \hspace{1em} \texttt{(SL:DESTINATION "work" )} \\
        &  & \hspace{1em} \texttt{\textcolor{dkgreen}{\textbf{(SL:OBSTRUCTION}} "traffic" ) )} \vspace{.5em} \\
        \multicolumn{3}{l}{\textbf{Unchanged}} \vspace{.5em} \\
        & \multicolumn{2}{l}{Query: \emph{What is the best route to get to Atlanta to see my brother Mark?}} \vspace{.5em} \\
        & \hspace{2em} V1 Label: & \hspace{2em} V2 Label: \\
        & \texttt{(IN:GET\_DIRECTIONS} & \texttt{(IN:GET\_DIRECTIONS} \\
        & \hspace{1em} \texttt{(SL:DESTINATION "Atlanta" ) )} & \hspace{1em} \texttt{(SL:DESTINATION "Atlanta" ) )} \vspace{.5em} \\
        \multicolumn{3}{l}{\textbf{Trivially-unchanged}} \vspace{.5em} \\
        & \multicolumn{2}{l}{Examples not labeled with one of the seven intents that allow the \texttt{SL:OBSTRUCTION} slot are} \\
        & \multicolumn{2}{l}{trivially-unchanged.} \\
        
        \\ \bottomrule
    \end{tabular}
    \caption{Examples from different types of updates that we simulate.  To simulate updates, we use the original dataset as the V2 form, and modify some examples to a V1 form in a way that is not easily reversed.  We describe our data in terms of three partitions: \partition{changed}, \partition{unchanged}, and \partition{trivially-unchanged} -- as described in Section~\ref{sec:setup-updates}.  Note that while \partition{trivially-unchanged} examples can be easily identified from their V1 labels, \partition{changed} and \partition{unchanged} examples cannot be easily distinguished in V1.}
    \label{fig:examples}
\end{figure*}

\subsection{Creating Synthetic Changes}
\label{sec:synthetic-changes}

To enable exploration of the conflicting data problem, we demonstrate a method for easily simulating updates with conflicts on an existing publicly available dataset.
We form our synthetic updates in the reverse direction -- the data from the original dataset represents the final V2 form, and we modify some examples to represent a V1 (pre-update) form.
We aim to make changes that can be done automatically in the reverse direction, while still being interesting to learn in the forward direction.

To form our updates, we select a particular intent or argument from the dataset to simulate the introduction of.
We sample a set of examples labeled with the selected intent or argument to form the primary part of the V2 training set.
For other examples with this intent or argument, we will keep them as part of the V1 training set, but re-label them to some form they may have taken before the new intent or argument was introduced.
These re-labeled examples act as conflicting data.

For example, suppose we would like to simulate the creation of the intent \texttt{GET\_INFO\_TRAFFIC}, as shown at the top of Figure~\ref{fig:examples}.
After selecting a subset of \texttt{GET\_INFO\_TRAFFIC} examples for the V2 training set, we form a conflicting set in the V1 data by re-labeling \texttt{GET\_INFO\_TRAFFIC} examples to use the related intent \texttt{GET\_INFO\_ROAD\_CONDITION}.
For this particular change, we kept the arguments the same when updating, but for others we remove arguments or relabel them in the V1 set.

Because the examples with our ``new'' intent or argument are different between the V1 and V2 data, those examples make up the \partition{changed} subset.  Recall that we would also like to include some \partition{unchanged} examples in the V2 training set.  To select examples for this partition, we find other examples labeled with the intent that we re-labeled our conflicting set to have.  For our \texttt{GET\_INFO\_TRAFFIC} intent example, this means finding some examples with the label \texttt{GET\_INFO\_ROAD\_CONDITION}, since these examples look similar to the re-labeled conflicting examples but should have the same labels in V1 and V2.
Since all of the examples where this update applies are labeled as \texttt{GET\_INFO\_ROAD\_CONDITION} in the V1 training set, any example with a different V1 label can be considered \partition{trivially-unchanged}.

In total, we form five different synthetic updates to use in our experiments.\footnote{Data for our synthesized updates can be downloaded from \url{https://github.com/google/overcoming-conflicting-data/}, and Appendix~\ref{sec:partition-sizes} summarizes the partition sizes for each of these changes.}
Figure~\ref{fig:examples} also shows examples for two other types of updates: introducing a new intent for previously unsupported inputs and introducing a new argument.
The updates are formed from the TOP dataset, which contains over 40,000 English queries about navigation or events labeled with tree-based semantic parses \cite{gupta-etal-2018-semantic-parsing}.

\section{Base Model}
\label{sec:base-model}

The semantic parsing model we use for our experiments is a sequence-to-sequence model based on the transformer architecture \cite{vaswani2017attention}.
We use a sequence-to-sequence model because of their flexibility and widespread use in semantic parsing \cite{jia-liang-2016-data,dong-lapata-2016-language,rongali2020dontparse}  and NLP in general.
Our model encodes a language input using a pre-trained 12-layer BERT model \cite{devlin-etal-2019-bert}, then decodes a parse tree flattened by depth-first traversal.  At each step, the decoder can generate either 1) a labeled bracket representing an intent or argument label 2) a closing bracket or 3) an index of an input token to be copied.
The hyperparameters of our model architecture and training can be found in Appendix~\ref{sec:model-hparams}.

\section{Effect of Conflicting Data}
\label{sec:conflicting-test}

Before we describe and test our methods for mitigating conflicts, this section will briefly explore how conflicting data affects learning.
We evaluate model updates both with and without conflicting data, and compare accuracies as we vary the amount of new V2 data being introduced.
The non-conflicting setting represents an oracle where all conflicting data is removed, which we can easily simulate in our synthetic data-creation process but is generally not achievable on real-world data without additional manual annotation.

More precisely, when evaluating updates with conflicting data, we include 50 examples with out-of-date labels in the \partition{changed} category.  We mix these examples with a full set of \partition{unchanged} and \partition{trivially-unchanged} examples to represent a V1 training set from before an update.
We then introduce different amounts of \partition{changed} examples with updated labels to act as the V2 training set, with sizes ranging from from half the conflicting set size (25) to four times as much (200).
For each data size, we measure accuracy using an exact match metric, meaning the entire tree output by the model must match the reference to be considered correct.
For the non-conflicting setting, we do not include the 50 examples with out-of-date labels, but otherwise use the same setup.

\begin{figure}
    \centering
    \includegraphics[width=\columnwidth]{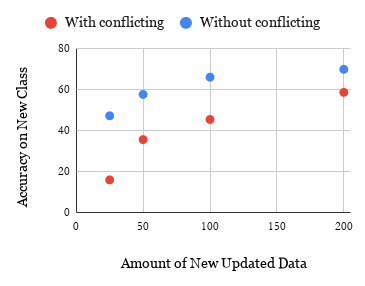}
    \caption{Accuracy as a function of data size with conflicting data compared to accuracy when an oracle removes the conflicts, averaged across five different updates.
    }
    \label{fig:conflict_test}
\end{figure}

The results for this experiment are shown in Figure~\ref{fig:conflict_test}, after averaging over five different changes (a detailed breakdown of results can be found in Appendix~\ref{sec:conflict-result-detail}).
To the left of the graph, we see that when the amount of conflicting data is greater than the size of the new data being added, we get less than half of the accuracy we would get without conflicting data.
While the gap narrows somewhat with more data, even when we introduce four times as much new data as there are conflicting examples, the presence of conflicting data still leads to a loss of over 10\% accuracy.
These results show just how detrimental conflicting data can be to the learning of a model update.

\section{Mitigation Methods}
\label{sec:methods}

In this work, we consider three methods to alleviate the problems caused by conflicting data: fine-tuning, multi-task learning with separate decoder heads, and data filtering with a learned classifier.

\subsection{Fine-tuning}

Our first and simplest method for handling conflicting data is fine-tuning.
For this approach, we first train a model on only the V1 training data, then after training completes, we take the final parameters and use those as initialization for training with the V2 data.  This approach can alleviate the conflicting data problem because the second stage of training does not include any conflicting data and the model will have less confusion about how to label \partition{changed} examples.  By first training on the V1 data, we are also able to benefit indirectly from the larger amount of data contained in it.

During the second stage of training, we train on the \partition{changed} and \partition{unchanged} data in the V2 training set, as well as the \partition{trivially-unchanged} examples from the V1 training set.  \partition{Trivially-unchanged} data from the V1 data can be included in V2 training because it can be easily identified and is known to not conflict.

\subsection{Multi-task}

\begin{figure}
    \centering
    \includegraphics[width=\columnwidth]{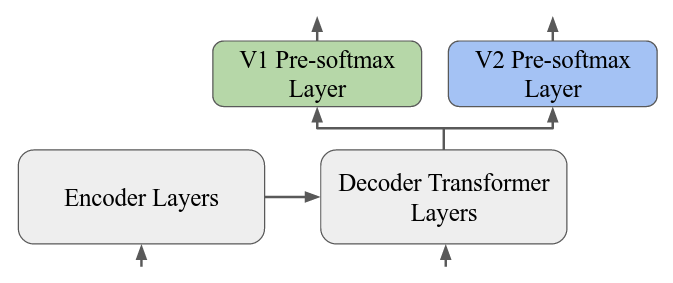}
    \caption{In our multi-task method, we feed the new V2 data to a separate decoder head to separate it from the possibly-conflicting data in the original V1 training set.  The encoder layers and decoder transformer layers are shared between V1 and V2.
    }
    \label{fig:multi-task}
\end{figure}

For our next method, we use an approach from multi-task learning where multiple decoder heads are used for different sets of data \cite[][]{caruana1997multitask,fan-etal-2017-transfer}.
Our two ``tasks'' correspond to the V1 data and the V2 data, and we use a separate set of parameters for the final pre-softmax layer of the decoder for each of the versions (as illustrated in Figure~\ref{fig:multi-task}).
The V1 head is only trained with V1 data and the V2 head is only trained with V2 data\mpavan{does the V2 head also include trivially-changed slice from V1}, but the encoder layers and decoder transformer are shared between both.
This way, the V2 head is never trained on conflicting examples from V1, but the overall V2 model can still benefit from some information in V1 data indirectly through the shared encoder and decoder transformer layers.  After training, the V1 head can be discarded, and the V2 head is used to make decisions.
We train on both versions simultaneously, with each batch containing some of both types of data.  As with our other methods, the V2 head is also trained on \partition{trivially-unchanged} data from the V1 data.

The goal behind this approach is similar to that of fine-tuning: to avoid training the V2 model directly on the possibly-conflicting V1 data while still sharing some amount of information through model parameters.
However, unlike fine-tuning, which is liable to forget information from V1 as training progresses, the simultaneous training for the multi-task method keeps the V1 information active for better sharing.

\subsection{Classifier-based Data Selection}
\label{sec:selection}

\begin{figure}
    \centering
    \includegraphics[width=.75\columnwidth]{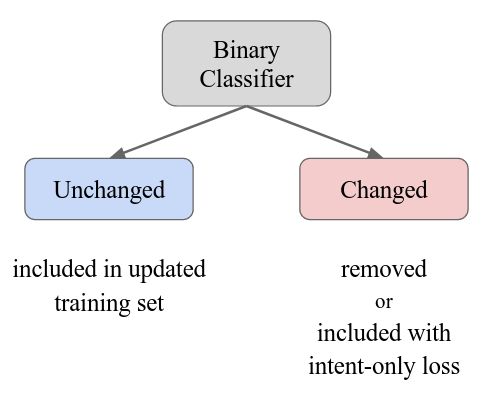}
    \caption{Our selection classification method uses a binary classifier to filter the original V1 data based on whether it is likely to be out-of-date.}
    \label{fig:selection}
\end{figure}

The final method we explore in this work is classifier-based data selection.
The idea behind the data selection strategy is to explicitly select examples from the original V1 data that we don’t think will conflict.
Using the small amount of V2 data, we train a classifier to predict whether an example will be \partition{changed} or \partition{unchanged}, and then apply this classifier to the V1 data, as illustrated in Figure~\ref{fig:selection}.
We can then include the selected examples in our updated training set, allowing us to take advantage of more information from the original training set while filtering out many of the problematic conflicting examples.

We first train a classifier on the V2 data to learn a binary decision between \partition{changed} and \partition{unchanged} examples.  This training requires that we can distinguish between which examples are \partition{changed} and \partition{unchanged}, which can either be specified as part of the annotation process, or can be estimated by running an existing V1 model to predict old labels for the provided V2 examples (in this work we use the annotation method, as part of our synthetic data-creation process).
Our classifier uses the same BERT encoder as our sequence-to-sequence parsing model and is initialized with parameters from a parsing model for the V1 data.
Representations of this encoder are averaged across time before feeding into a small feedforward network with a hidden dimension of 512.
After training on the V2 set, the binary classifier is run on the V1 training set (excluding the \partition{trivially-unchanged} examples, which can be automatically included as-is for V2 training).  This creates a categorization of predicted-changed and predicted-unchanged, which we hope will closely approximate the true \partition{changed} and \partition{unchanged} sets.  For many of the changes we tested, the classifiers performed quite well, with accuracies above 90\% on held-out data.  For examples in predicted-unchanged, we will include them in the set of training examples used to train our updated model.  For examples in predicted-changed, we do not want to directly include them, and consider two possible solutions: 1) remove them completely or 2) include the examples with an intent-only loss, as described below.

\subsubsection{Intent-only Loss}

The idea of the intent-only loss variant of the data selection method is to try to take advantage of more information about the predicted-changed examples without requiring us to know the full form of the tree after a change.  While we know that these examples are likely to have changed, in general we do not know if or how the full argument structure will change for all examples.  However, it is usually possible to know what the intent should be for the \partition{changed} examples.  If the update is introducing a new intent, we can use this new intent for the predicted-changed examples.  If the update only affects arguments, we can keep the original intents for the examples.
Which of these cases applies can be specified manually (as we do in our experiments), or could likely be determined automatically by running a V1 model on the V2 data and comparing the intents.
Once we have determined the new top-level intent for the predicted-changed examples, we will include them as special training examples that only receive a loss on their intent.  Since the intent is the first token to be predicted by the sequence-to-sequence decoder, we can simply mask out the loss for the rest of the tokens in the sequence.  With this masking, argument structure prediction will be unaffected by these examples, and the model must defer to other examples, such as those in the V2 training set, to learn argument labeling.

One case that this approach does not currently handle are updates that introduce multiple new intents simultaneously, and we leave an exploration of that case to future work.
To use an intent-only loss on those updates, more fine-grained classifications are needed to determine what the correct intent for the \partition{changed} examples are.

\section{Evaluation}
\label{sec:evaluation}

\begin{table}
    \centering
    \begin{tabular}{l r}
        \toprule
         & \textbf{Overall Accuracy}  \\ \midrule
        \textbf{Baselines} \\
        \quad Train on V1 data & 49.2 \\
        \quad Train on V2 data & 51.8 \\
        \quad Direct mixing & 49.9 \\
        \midrule
        \textbf{Methods} \\
        \quad Fine-tune & 59.9 \\
        \quad Multi-task & 69.7 \\
        \quad Selection (drop changed) & 65.6 \\
        \quad Selection (intent only) & \textbf{71.5} \\
        \midrule 
        \textbf{Oracle} \\
        \quad Relabel all examples & 74.7 \\
        \bottomrule
    \end{tabular}
    \caption{Exact match accuracies of our methods for mitigating the effects of conflicting data, averaged across five different updates.  Our methods greatly outperform the baselines and close a substantial portion of the gap to the oracle upper bound.}
    \label{tab:results}
\end{table}

\begin{figure*}[t]
    \centering
    \includegraphics[width=.9\textwidth]{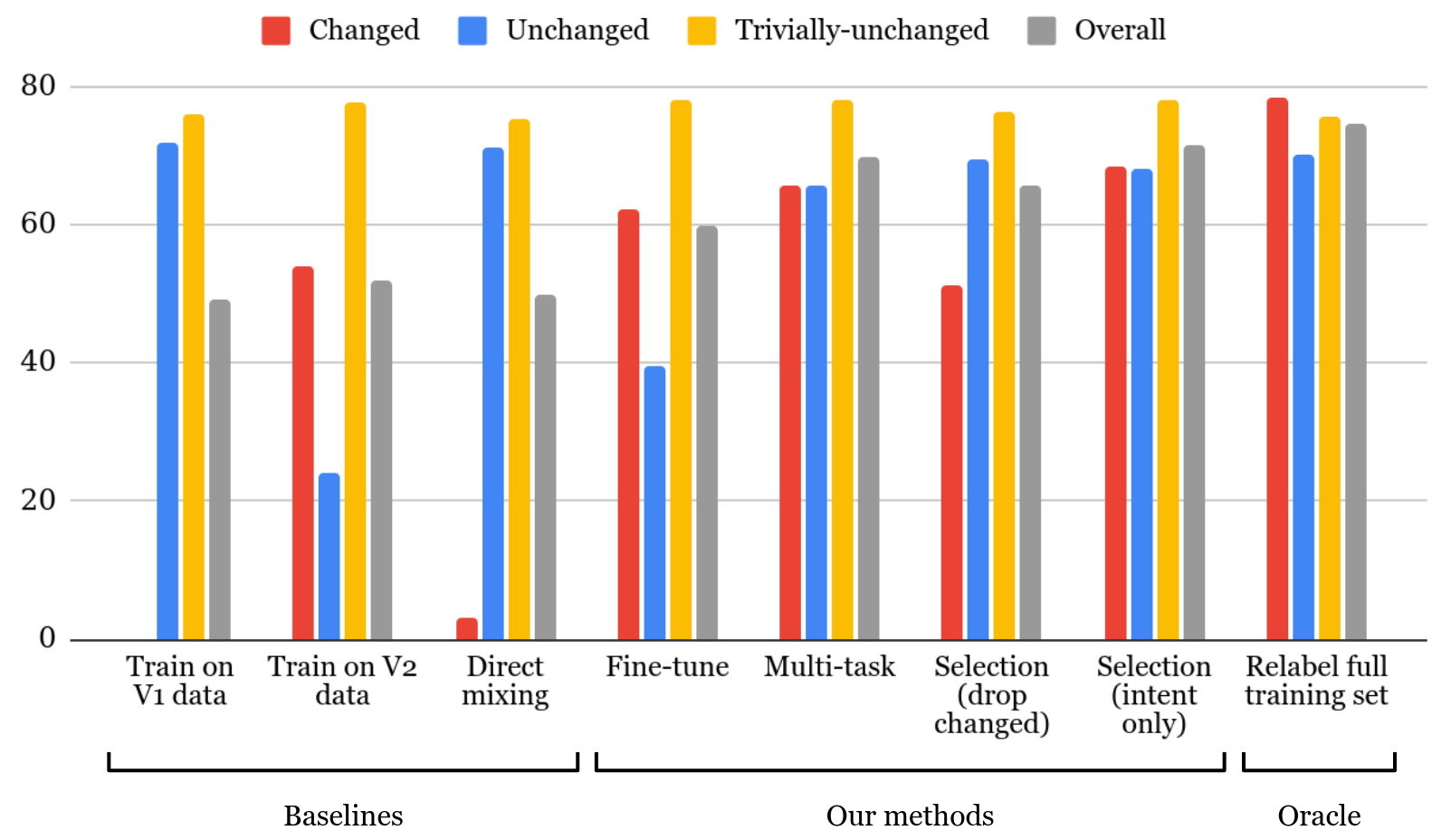}
    \caption{A break down of results across the three data partitions.
    }
    \label{fig:results_chart}
\end{figure*}

In this section, we describe the evaluation of our methods for mitigating the conflicting data problem.
For each update, we form a test set by randomly selecting 100 examples from each of the three data partitions (\partition{changed}, \partition{unchanged}, and \partition{trivially-unchanged}), and a V2 training set by selecting 50 \partition{changed} and 50 \partition{unchanged} examples.
All remaining examples are placed into the V1 training set and \partition{changed} examples are relabeled appropriately.  Note that unlike in Section~\ref{sec:conflicting-test} where we used a fixed-size  conflicting set of 50 examples in V1, in this setup we use all remaining examples available after sampling a subset for V2, which results in larger conflicting sets ranging from hundreds to thousands of examples.  The large sizes of the conflicting sets further amplifies the effect of the conflicting data.
We report results in terms of exact match accuracy between the predicted and target parse structure.
Results are aggregated across 5 different updates, giving us a total of 1500 test examples for each method (5 updates $\times$ 3 partitions $\times$ 100 examples).  We also average across 5 different runs for each method to reduce variance.

We compare against three baselines: training only on the original V1 data, training only on the new V2 data, and directly mixing the two data versions together into a single dataset.\footnote{We also tried a variant of the direct mixing baseline where the V2 data is upsampled to try to account for differences in size, but this obtained almost identical results, indicating that upsampling is not an effective method for overcoming conflicting data (not shown in main results; see Appendix~\ref{sec:main-result-detail}).}
We also compare to an upper bound where the entire V1 training set is re-annotated with updated labels.  For many updates, this upper bound requires thousands of new annotations, as compared to the one hundred labels used by our methods.

\subsection{Results}

The results of our evaluation are shown in Table~\ref{tab:results}.  All of our methods substantially outperform the baselines.  Our best method, the selection classifier with an intent-only loss on changed examples (\S~\ref{sec:selection}), obtains an accuracy of 71.5\%, covering 86\% of the gap between the best baseline and the oracle upper bound.

To see a more clear picture of what is happening, we also break down results by data partitions, as shown in Figure~\ref{fig:results_chart} (an even more detailed breakdown across different updates is provided in Appendix~\ref{sec:main-result-detail}).
In this chart, we can see that the baseline that mixes the V1 and V2 data without accounting for conflicts performs extremely poorly on the \partition{changed} examples, echoing our results in Section~\ref{sec:conflicting-test}.
On the other hand, training only on the small set of V2 data throws away all the information from the original V1 training set, limiting its performance (particularly on \partition{unchanged} examples).
Our methods provide an effective way to combine information in both datasets without overwhelming \partition{changed} data with out-of-date labels.

\section{Conclusion}

This work has shown that in order to make effective updates to the outputs of a neural semantic parsing model by adding new data, it is important to consider the effect of conflicting examples in the original data.
Conflicting data is likely a problem in many different scenarios where outputs to a model must be updated, and we believe that further study into methods for mitigating the effects of conflicting data is an important direction to allow practitioners to handle the constantly-changing needs of real-world machine learning applications.

\bibliographystyle{acl_natbib}
\bibliography{references}

\begin{thebibliography}{19}
\expandafter\ifx\csname natexlab\endcsname\relax\def\natexlab#1{#1}\fi

\bibitem[{Caruana(1997)}]{caruana1997multitask}
Rich Caruana. 1997.
\newblock Multitask learning.
\newblock \emph{Machine learning}, 28(1):41--75.

\bibitem[{Castro et~al.(2018)Castro, Mar{\'\i}n-Jim{\'e}nez, Guil, Schmid, and
  Alahari}]{castro2018end}
Francisco~M Castro, Manuel~J Mar{\'\i}n-Jim{\'e}nez, Nicol{\'a}s Guil, Cordelia
  Schmid, and Karteek Alahari. 2018.
\newblock End-to-end incremental learning.
\newblock In \emph{Proceedings of the European conference on computer vision
  (ECCV)}, pages 233--248.

\bibitem[{Chen and Moschitti(2019)}]{chen2019transfer}
Lingzhen Chen and Alessandro Moschitti. 2019.
\newblock Transfer learning for sequence labeling using source model and target
  data.
\newblock In \emph{Proceedings of the AAAI Conference on Artificial
  Intelligence}, volume~33, pages 6260--6267.

\bibitem[{Devlin et~al.(2019)Devlin, Chang, Lee, and
  Toutanova}]{devlin-etal-2019-bert}
Jacob Devlin, Ming-Wei Chang, Kenton Lee, and Kristina Toutanova. 2019.
\newblock \href {https://doi.org/10.18653/v1/N19-1423} {{BERT}: Pre-training of
  deep bidirectional transformers for language understanding}.
\newblock In \emph{Proceedings of the 2019 Conference of the North {A}merican
  Chapter of the Association for Computational Linguistics: Human Language
  Technologies, Volume 1 (Long and Short Papers)}, pages 4171--4186,
  Minneapolis, Minnesota. Association for Computational Linguistics.

\bibitem[{Dong and Lapata(2016)}]{dong-lapata-2016-language}
Li~Dong and Mirella Lapata. 2016.
\newblock \href {https://doi.org/10.18653/v1/P16-1004} {Language to logical
  form with neural attention}.
\newblock In \emph{Proceedings of the 54th Annual Meeting of the Association
  for Computational Linguistics (Volume 1: Long Papers)}, pages 33--43, Berlin,
  Germany. Association for Computational Linguistics.

\bibitem[{Fan et~al.(2017)Fan, Monti, Mathias, and
  Dreyer}]{fan-etal-2017-transfer}
Xing Fan, Emilio Monti, Lambert Mathias, and Markus Dreyer. 2017.
\newblock \href {https://doi.org/10.18653/v1/W17-2607} {Transfer learning for
  neural semantic parsing}.
\newblock In \emph{Proceedings of the 2nd Workshop on Representation Learning
  for {NLP}}, pages 48--56, Vancouver, Canada. Association for Computational
  Linguistics.

\bibitem[{Gupta et~al.(2018)Gupta, Shah, Mohit, Kumar, and
  Lewis}]{gupta-etal-2018-semantic-parsing}
Sonal Gupta, Rushin Shah, Mrinal Mohit, Anuj Kumar, and Mike Lewis. 2018.
\newblock \href {https://doi.org/10.18653/v1/D18-1300} {Semantic parsing for
  task oriented dialog using hierarchical representations}.
\newblock In \emph{Proceedings of the 2018 Conference on Empirical Methods in
  Natural Language Processing}, pages 2787--2792, Brussels, Belgium.
  Association for Computational Linguistics.

\bibitem[{Jia and Liang(2016)}]{jia-liang-2016-data}
Robin Jia and Percy Liang. 2016.
\newblock \href {https://doi.org/10.18653/v1/P16-1002} {Data recombination for
  neural semantic parsing}.
\newblock In \emph{Proceedings of the 54th Annual Meeting of the Association
  for Computational Linguistics (Volume 1: Long Papers)}, pages 12--22, Berlin,
  Germany. Association for Computational Linguistics.

\bibitem[{Jiang et~al.(2018)Jiang, Zhou, Leung, Li, and
  Fei-Fei}]{Jiang2018MentorNetLD}
Lu~Jiang, Zhengyuan Zhou, T.~Leung, L.~Li, and Li~Fei-Fei. 2018.
\newblock Mentornet: Learning data-driven curriculum for very deep neural
  networks on corrupted labels.
\newblock In \emph{ICML}.

\bibitem[{Kirkpatrick et~al.(2017)Kirkpatrick, Pascanu, Rabinowitz, Veness,
  Desjardins, Rusu, Milan, Quan, Ramalho, Grabska-Barwinska
  et~al.}]{kirkpatrick2017overcoming}
James Kirkpatrick, Razvan Pascanu, Neil Rabinowitz, Joel Veness, Guillaume
  Desjardins, Andrei~A Rusu, Kieran Milan, John Quan, Tiago Ramalho, Agnieszka
  Grabska-Barwinska, et~al. 2017.
\newblock Overcoming catastrophic forgetting in neural networks.
\newblock \emph{Proceedings of the national academy of sciences},
  114(13):3521--3526.

\bibitem[{Li and Hoiem(2016)}]{li2016learning}
Zhizhong Li and Derek Hoiem. 2016.
\newblock Learning without forgetting.
\newblock In \emph{European Conference on Computer Vision}, pages 614--629.
  Springer.

\bibitem[{Lu et~al.(2018)Lu, Liu, Dong, Gu, Gama, and Zhang}]{lu2018learning}
Jie Lu, Anjin Liu, Fan Dong, Feng Gu, Joao Gama, and Guangquan Zhang. 2018.
\newblock Learning under concept drift: A review.
\newblock \emph{IEEE Transactions on Knowledge and Data Engineering},
  31(12):2346--2363.

\bibitem[{Rongali et~al.(2020)Rongali, Soldaini, Monti, and
  Hamza}]{rongali2020dontparse}
Subendhu Rongali, Luca Soldaini, Emilio Monti, and Wael Hamza. 2020.
\newblock \href {https://doi.org/10.1145/3366423.3380064} {Don’t parse,
  generate! a sequence to sequence architecture for task-oriented semantic
  parsing}.
\newblock In \emph{Proceedings of The Web Conference 2020}, WWW '20, page
  2962–2968, New York, NY, USA. Association for Computing Machinery.

\bibitem[{Rusu et~al.(2016)Rusu, Rabinowitz, Desjardins, Soyer, Kirkpatrick,
  Kavukcuoglu, Pascanu, and Hadsell}]{rusu2016progressive}
Andrei~A Rusu, Neil~C Rabinowitz, Guillaume Desjardins, Hubert Soyer, James
  Kirkpatrick, Koray Kavukcuoglu, Razvan Pascanu, and Raia Hadsell. 2016.
\newblock Progressive neural networks.
\newblock \emph{arXiv preprint arXiv:1606.04671}.

\bibitem[{Sukhbaatar et~al.(2014)Sukhbaatar, Bruna, Paluri, Bourdev, and
  Fergus}]{sukhbaatar2014training}
Sainbayar Sukhbaatar, Joan Bruna, Manohar Paluri, Lubomir Bourdev, and Rob
  Fergus. 2014.
\newblock Training convolutional networks with noisy labels.
\newblock \emph{arXiv preprint arXiv:1406.2080}.

\bibitem[{Tsymbal(2004)}]{tsymbal2004problem}
Alexey Tsymbal. 2004.
\newblock The problem of concept drift: definitions and related work.
\newblock \emph{Computer Science Department, Trinity College Dublin},
  106(2):58.

\bibitem[{Vaswani et~al.(2017)Vaswani, Shazeer, Parmar, Uszkoreit, Jones,
  Gomez, Kaiser, and Polosukhin}]{vaswani2017attention}
Ashish Vaswani, Noam Shazeer, Niki Parmar, Jakob Uszkoreit, Llion Jones,
  Aidan~N Gomez, {\L}ukasz Kaiser, and Illia Polosukhin. 2017.
\newblock Attention is all you need.
\newblock In \emph{Advances in neural information processing systems}, pages
  5998--6008.

\bibitem[{Veit et~al.(2017)Veit, Alldrin, Chechik, Krasin, Gupta, and
  Belongie}]{Veit2017LearningFN}
Andreas Veit, Neil Alldrin, Gal Chechik, Ivan Krasin, A.~Gupta, and Serge~J.
  Belongie. 2017.
\newblock Learning from noisy large-scale datasets with minimal supervision.
\newblock \emph{2017 IEEE Conference on Computer Vision and Pattern Recognition
  (CVPR)}, pages 6575--6583.

\bibitem[{Xiao et~al.(2014)Xiao, Zhang, Yang, Peng, and Zhang}]{xiao2014error}
Tianjun Xiao, Jiaxing Zhang, Kuiyuan Yang, Yuxin Peng, and Zheng Zhang. 2014.
\newblock Error-driven incremental learning in deep convolutional neural
  network for large-scale image classification.
\newblock In \emph{Proceedings of the 22nd ACM international conference on
  Multimedia}, pages 177--186.

\end{thebibliography}

\clearpage
\pagebreak

\appendix
\onecolumn

\section{Model Hyperparameters}
\label{sec:model-hparams}

\begin{center}
\begin{tabular}{lr}
    \toprule
    \textbf{Hyperparameter} & \textbf{Value} \\ \midrule
    Decoder layers & 1 \\
    Decoder dimension & 256 \\
    Decoder feedforward dimension & 256 \\
    Batch size & 512 \\
    Training steps & 50000 \\
    Learning rate & 3e-4 \\
    Learning rate warmup & 10000 \\
    \bottomrule
\end{tabular}
\end{center}

These hyperparameters were kept constant across all experiments and were selected based on defaults from an existing implementation.
Model parameter counts are dominated by the BERT encoder, with approximately 100 million parameters.
Training was performed on TPUs and took several hours per run.

\section{List of changes with data sizes}
\label{sec:partition-sizes}
The table below briefly describes the five updates we test on with the sizes of each data partition \partition{changed}, \partition{unchanged}, and \partition{trivially-unchanged}.  For our primary experiments, 100 examples from each partition are placed in the test set, 50 examples from \partition{changed} and \partition{unchanged} are placed in the V2 training set, and the remainder are used for the V1 training set.

\begin{center}
    \begin{tabular}{l r r r}
        \toprule
        \textbf{Update type} & \textbf{Changed} & \textbf{Unchanged} & \textbf{Trivially-unchanged} \\ \midrule
        New intent from unsupported & 1719 & 1756 & 32266 \\
        New intent from related + argument name change & 3625 & 6776 & 25340 \\ 
        New argument & 635 & 21114 & 13992 \\
        New intent from related & 10044 & 422 & 25275 \\
        New intent from multiple intents in V1 & 285 & 3942 & 31514 \\
        \bottomrule
    \end{tabular}
\end{center}

\pagebreak
\section{Effect of Conflicting Data Results Detail}
\label{sec:conflict-result-detail}

The table below details the results for the experiment described in Section~\ref{sec:conflicting-test} and summarized in Figure~\ref{fig:conflict_test}.  We vary the size of updated data in the V2 \partition{changed} partition while holding constant a set of 50 conflicting examples in the original V1 data.

\begin{center}
    \begin{tabular}{l r r r r}
    \toprule
\textbf{Change} & \textbf{New data size} & \multicolumn{2}{c}{\textbf{Accuracy (\%)}} \\ \cmidrule{3-4}
& & \textbf{Conflicting} & \textbf{No Conflicting} \\ \midrule
New from unsupported & 25 & 21 & 42 \\
 & 50 & 27 & 51 \\
 & 100 & 54 & 60 \\
 & 200 & 61 & 71 \\
 \midrule
New from related + arg change & 25 & 17 & 59 \\
 & 50 & 44 & 73 \\
 & 100 & 52 & 74 \\
 & 200 & 66 & 75 \\
 \midrule
New argument & 25 & 11 & 54 \\
 & 50 & 44 & 54 \\
 & 100 & 45 & 68 \\
 & 200 & 67 & 67 \\
 \midrule
New from related & 25 & 16 & 41 \\
 & 50 & 29 & 56 \\
 & 100 & 42 & 56 \\
 & 200 & 54 & 71 \\
 \midrule
New from multiple sources & 25 & 15 & 40 \\
 & 50 & 34 & 54 \\
 & 100 & 34 & 72 \\
 & 200 & 45 & 65 \\
 \midrule
Average & 25 & 16 & 47.2 \\
 & 50 & 35.6 & 57.6 \\
 & 100 & 45.4 & 66 \\
 & 200 & 58.6 & 69.8 \\
  \bottomrule
    \end{tabular}
\end{center}

\pagebreak
\section{Main Results Detail}
\label{sec:main-result-detail}

The following table breaks down our main results across the different updates tested.  These results are described in Section~\ref{sec:evaluation} and summarized in Figure~\ref{fig:results_chart}.

\begin{center}
{\small
    \begin{tabular}{l l r r r r r r r r r r}
    \toprule
\textbf{Update} & \textbf{Partition} & \multicolumn{9}{c}{\textbf{Accuracy (\%)}} \\ \cmidrule{3-11}
& & \textbf{V1 only} & \textbf{V2 only} & \textbf{Dir mix} & \textbf{Upsampl} & \textbf{Fine-tu} & \textbf{Multi-t} & \textbf{Sel (rm)} & \textbf{Sel (io)} & \textbf{Oracle} \\ \midrule
A & Change & 0.0 & 53.6 & 0.8 & 0.0 & 58.2 & 66.8 & 51.4 & 68.2 & 80.2 \\
& Unchange & 75.4 & 20.8 & 70.0 & 74.8 & 35.0 & 61.6 & 72.8 & 69.8 & 72.2 \\
& Triv-unch & 78.8 & 77.2 & 77.8 & 77.6 & 78.8 & 82.0 & 78.0 & 79.4 & 78.0 \\

B & Change & 0.0 & 59.6 & 0.0 & 0.6 & 66.2 & 70.6 & 60.2 & 64.6 & 81.8 \\
& Unchange & 68.4 & 15.2 & 69.3 & 66.6 & 35.2 & 62.2 & 66.6 & 65.2 & 67.0 \\
& Triv-unch & 79.6 & 83.0 & 79.3 & 80.0 & 82.8 & 81.2 & 80.0 & 81.8 & 77.2 \\

C & Change & 0.0 & 31.2 & 1.8 & 2.8 & 45.4 & 39.0 & 43.8 & 59.2 & 73.4 \\
& Unchange & 79.0 & 25.4 & 81.0 & 80.6 & 42.4 & 79.2 & 77.2 & 80.8 & 81.4 \\
& Triv-unch & 73.4 & 75.0 & 72.3 & 73.6 & 74.2 & 74.8 & 72.4 & 74.6 & 73.6 \\

D & Change & 0.0 & 55.0 & 0.0 & 0.0 & 69.0 & 85.6 & 55.2 & 83.4 & 86.8 \\
& Unchange & 64.6 & 41.4 & 66.2 & 64.8 & 52.2 & 61.6 & 60.8 & 56.8 & 59.0 \\
& Triv-unch & 70.0 & 74.4 & 70.4 & 73.0 & 73.2 & 71.8 & 71.8 & 73.0 & 72.0 \\

E & Change & 0.0 & 69.6 & 13.0 & 13.4 & 71.8 & 65.4 & 46.2 & 66.0 & 69.8 \\
& Unchang & 71.0 & 17.4 & 69.8 & 72.2 & 33.4 & 63.0 & 69.0 & 68.2 & 70.6 \\
& Triv-unch & 78.2 & 78.6 & 77.3 & 77.6 & 80.8 & 80.2 & 79.2 & 81.2 & 77.2 \\ \midrule

Avg & Change & 0.0 & 53.8 & 3.1 & 3.4 & 62.1 & 65.5 & 51.4 & 68.3 & 78.4 \\
& Unchange & 71.7 & 24.0 & 71.2 & 71.8 & 39.6 & 65.5 & 69.3 & 68.2 & 70.0 \\
& Triv-unch & 76.0 & 77.6 & 75.4 & 76.4 & 78.0 & 78.0 & 76.3 & 78.0 & 75.6
 \\
 \bottomrule
    \end{tabular}
}
\end{center}
\textbf{Update key:} A: New intent from unsupported, B: New intent from related with argument relabeling, C: New argument, D: New intent from related with same arguments, E: New intent from multiple intents in V1
\end{document}